\def\year{2022}\relax
\documentclass[letterpaper]{article} 
\usepackage{aaai22}  
\usepackage{times}  
\usepackage{helvet}  
\usepackage{courier}  
\usepackage[hyphens]{url}  
\usepackage{graphicx} 
\usepackage{natbib}  
\usepackage{caption} 
\DeclareCaptionStyle{ruled}{labelfont=normalfont,labelsep=colon,strut=off} 
\usepackage{algorithm}
\usepackage{algorithmic}

\usepackage{booktabs} 
\usepackage{newfloat}
\usepackage{listings}
\floatstyle{ruled}
\newfloat{listing}{tb}{lst}{}
\floatname{listing}{Listing}
\usepackage{amsmath,amsfonts,bm}
\usepackage{amsmath}

\def\eqref#1{equation~\ref{#1}}

\def\1{\bm{1}}

\def\vw{{\bm{w}}}
\def\vx{{\bm{x}}}

\def\mC{{\bm{C}}}

\def\mI{{\bm{I}}}

\def\mS{{\bm{S}}}

\def\mW{{\bm{W}}}

\DeclareMathAlphabet{\mathsfit}{\encodingdefault}{\sfdefault}{m}{sl}
\SetMathAlphabet{\mathsfit}{bold}{\encodingdefault}{\sfdefault}{bx}{n}
\newcommand{\tens}[1]{\bm{\mathsfit{#1}}}

\def\tW{{\tens{W}}}

\title{WLD-Reg: A Data-dependent Within-layer Diversity Regularizer}
\author{
    Firas Laakom\textsuperscript{\rm 1}\thanks{This  work  was  supported  by   NSF-Business Finland  Center  for Visual and Decision Informatics (CVDI) project AMALIA.}
    Jenni Raitoharju,\textsuperscript{\rm 2}
   Alexandros Iosifidis,\textsuperscript{\rm 3}
    Moncef Gabbouj \textsuperscript{\rm 1}
}
\affiliations{
    \textsuperscript{\rm 1} Faculty of Information Technology and Communication Sciences, Tampere University,  Finland\\
    \textsuperscript{\rm 2} Faculty of Information Technology, University of Jyväskylä, Finland \\

    \textsuperscript{\rm 3} DIGIT, Department of Electrical and Computer Engineering, Aarhus University, Denmark \\

}

\usepackage{bibentry}

\begin{document}

\maketitle

\begin{abstract}
Neural networks are composed of multiple layers arranged in a hierarchical structure jointly trained with a gradient-based optimization, where the errors are back-propagated from the last layer back to the first one. At each optimization step, neurons at a given layer receive feedback from neurons belonging to higher layers of the hierarchy. In this paper, we propose to complement this traditional 'between-layer' feedback with additional 'within-layer' feedback to encourage the diversity of the activations within the same layer. To this end, we measure the pairwise similarity between the outputs of the neurons and use it to model the layer's overall diversity. We present an extensive empirical study confirming that the proposed approach enhances the performance of several state-of-the-art neural network models in multiple tasks. The code is publically available at \url{https://github.com/firasl/AAAI-23-WLD-Reg}
\end{abstract}
\section{Introduction}
Deep learning has been extensively used in the last decade to solve several tasks \cite{krizhevsky2012imagenet,golan2018deep,hinton2012deep}.  A deep learning model, i.e., a neural network, is formed of a sequence of layers with parameters optimized during the training process using training data. 
Formally, an m-layer neural network  model can be defined as follows:
\begin{equation} 
f(\vx;\tW) = \phi^m(\mW^m(\phi^{m-1}( \cdots \phi^2(\mW^2 \phi^1(\mW^1 \vx) ))), 
\end{equation}
where $\phi^i(.)$ is the non-linear activation function of the $i^{th}$ layer and $\tW= \{\mW^1,\dots,\mW^m\}$ are the model's weights. Given a training data $\{\vx_i,y_i\}_{i=1}^{N}$, the parameters of $f(\vx;\tW)$ are obtained by minimizing a loss $\hat{L}(\cdot)$:
\begin{equation} \label{eq:loss}
\hat{L}(f) = \frac{1}{N} \sum_{i=1}^{N} l\big(f(\vx_i;\tW),y_i \big).
\end{equation}

However,  neural networks are often over-parameterized, i.e., have more parameters than data. As a result, they tend to overfit to the training samples and not generalize well on  unseen examples \cite{goodfellow2016deep}. While research on double descent \cite{advani2020high,belkin2019reconciling,Nakkiran2020Deep} shows that over-parameterization does not necessarily lead to overfitting, avoiding overfitting has been extensively studied \cite{dziugaite2017computing,foret2020sharpness,nagarajan2019uniform,neyshabur2018role,poggio2017theory,grari2021learning} and  various approaches and strategies, such as data augmentation \cite{goodfellow2016deep,zhang2018mixup},  regularization \cite{arora2019implicit,bietti2019kernel,kukavcka2017regularization,ouali2021spatial,han2021continual}, and Dropout \cite{hinton2012improving,lee2019meta,li2016improved,wang2019rademacher}, have been proposed to close the gap between the empirical loss and the expected loss.

Diversity of learners is widely known to be important in ensemble learning \cite{li2012diversity,yu2011diversity} and, particularly in deep learning context, diversity of  information extracted by the network neurons has been recognized as a viable way to improve generalization \cite{xie2017diverse,xie2015generalization}. In most cases, these efforts have focused on making the set of weights more diverse \cite{yang2019balanced,malkin2009multi}. However,  diversity of the activations has not received much attention. Here, we argue that due to the presence of non-linear activations, diverse weights do not guarantee diverse feature representation. Thus, we propose focusing on the diversity on top of feature mapping instead of the weights.

To the best of our knowledge, only \cite{cogswell2015reducing,div_theory} have considered  diversity of the activations directly in the neural network context. The work in \cite{div_theory} studied theoretically how diversity affects generalization showing that it can reduce overfitting. The work in \cite{cogswell2015reducing} proposed an additional loss term using  cross-covariance of hidden activations, which encourages the neurons to learn diverse or non-redundant representations. The proposed approach, known as DeCov, was empirically proven to alleviate overfitting and to improve the generalization ability of neural network. However, modeling diversity as the sum of the pairwise cross-covariance, it is not scale-invariant and can lead to trivial solutions. Moreover, it can capture only the pairwise diversity between components and is unable to capture the "higher-order diversity”.

In this work, we propose a novel approach to encourage activation diversity within the same layer. We propose complementing the 'between-layer' feedback with additional 'within-layer' feedback to penalize similarities between  neurons on the same layer. Thus, we encourage each neuron to learn a distinctive representation and to enrich the data representation learned within each layer. We propose three variants for our approach that are based on different global diversity definitions.

Our contributions in this paper are as follows: 
\begin{itemize}
    \item We propose a new approach to encourage the 'diversification' of the layers' output feature maps in neural networks. The proposed approach has three variants. The main intuition is that, by promoting the within-layer activation diversity, neurons within a layer learn distinct patterns and, thus, increase the overall capacity of the model.  
                                                        
    \item We show empirically that the proposed within-layer activation diversification boosts the performance of neural networks. Experimental results on several tasks show that the proposed approach outperforms competing methods.
\end{itemize}

\section{Within-layer Diversity Regularizer} \label{our_approach}
 
In this section, we propose a novel diversification strategy, where we encourage neurons within a layer to activate in a mutually different manner, i.e., to capture different patterns. In this paper, we define as “feature layer” the last intermediate layer in a neural network. In the rest of the paper, we focus on this layer and propose a data-dependent regularizer which forces each unit within this layer to learn a distinct pattern and penalizes the similarities between the units. Intuitively, the proposed approach reduces the reliance of the model on a single pattern and, thus, can improve generalization. 

We start by modeling the global similarity between two units. Let $\phi_n(\vx_j)$ and $\phi_m(\vx_j)$ be the outputs of the $n^{th}$ and $m^{th}$ unit in the feature layer for the same input sample $\vx_j$. The similarity $s_{nm}$ between the the $n^{th}$ and $m^{th}$ neurons can be obtained as the average similarity measure of their outputs for $N$ input samples. We use the radial basis function to express the similarity:
\begin{equation} \label{eq:snm}
s_{nm} := \frac{1}{N} \sum_{j=1}^N  \text{exp} \big(-\gamma || \phi_n(\vx_j) - \phi_m(\vx_j)||\big),
\end{equation}
where $\gamma$ is a hyper-parameter. The similarity $s_{nm}$ can be computed over the whole dataset or batch-wise. Intuitively, if two neurons $n$ and $m$ have similar outputs for  many samples, their corresponding similarity $s_{nm}$ will be high. Otherwise, their similarity $s_{mn}$ is small and they are considered ``diverse".  

Next, based on these pairwise similarities, we propose three variants for obtaining the overall similarity $J$ of all the units within the feature layer: 
\begin{itemize}
    \item \textbf{Direct:}  $J :=\sum_{n\neq m} s_{nm} $. In this variant, we model the global layer similarity directly as the sum of the pairwise similarities between the neurons. By minimizing their sum, we encourage the neurons to learn different representations.
    \item \textbf{Det:}  $J := -\text{det}(\textbf{S})$, where $\mS$ is a similarity matrix defined as $\mS_{nm}=s_{nm}$. This variant is inspired by the Determinantal Point Process (DPP) \cite{kulesza2010structured,kulesza2012determinantal}, as the determinant of $\mS$ measures the global diversity of the set. Geometrically, $\text{det}(\mS)$ is the volume of the parallelepiped formed by vectors in the feature space associated with $s$. Vectors that result in a larger volume are considered to be more ``diverse".  Thus, maximizing $\text{det}(\cdot)$ (minimizing $-\text{det}(\cdot)$) encourages the diversity of the learned features. 
    \item \textbf{Logdet:}  $J := -\text{logdet}(\textbf{S})$\footnote{This is defined only if $\mS$ is positive definite. It can be shown that in our case $\mS$ is positive semi-definite. Thus, in practice, we use a regularized version ($\mS +\epsilon \mI$) to ensure the positive definiteness.}. This variant has the same motivation as the second one. We use Logdet instead of Det as Logdet is a convex function over the positive definite matrix space.  
\end{itemize}

It should be noted here that the first proposed variant, i.e., direct, similar to DeCov \cite{cogswell2015reducing}, captures only the pairwise similarity between components and is unable to capture the higher-order “diversity”, whereas the other two variants consider the global similarity and are able to measure diversity in a more global manner. Promoting diversity of activations within a layer can lead to tighter generalization bound and can theoretically decrease the gap between the empirical and the true risks \cite{div_theory}.

The proposed global similarity measures $J$ can be minimized by using them as an additional loss term. However, we note that the pair-wise similarity measure $s_{nm}$, expressed in \eqref{eq:snm}, is not scale-invariant. In fact, it can be trivially minimized by making all activations of the feature layer  high, i.e., by multiplying by a high scaling factor, which has no effect on the performance, since the model can rescale high activations to normal values simply by learning small weights on the next layer. To alleviate this problem, we propose an additional term, which penalizes high activation values. The total proposed additional loss is defined as follows: 
\begin{equation} \label{eq:lossextra}
 \hat{L}_{WLD-Reg}  := \lambda_1 J + \lambda_2 \sum_{i=1}^N || \Phi(\vx_i)||^2_2,
\end{equation}
where $\Phi(\vx) =[\phi_1(\vx), \cdots, \phi_C(\vx)]$ is the feature vector, $C$ is the number of units within the feature layer, and  $\lambda_1$ and $\lambda_2$ are two hyper-parameters controlling the contribution of each term to the diversity loss. Intuitively, the first term of \eqref{eq:lossextra} penalizes the similarity between the units and promotes diversity, whereas the second term ensures the scale-invariance of the proposed regularizer. 

The total loss function $\hat{L}(f)$ defined in \eqref{eq:loss} is augmented as follows: 
\begin{align} \label{eq:augloss}
\hat{L}_{aug}(f) & := \hat{L}(f) + \hat{L}_{WLD-Reg} \\
                 & = \hat{L}(f) + \lambda_1 J + \lambda_2 \sum_{i=1}^N || \Phi(\vx_i)||^2_2. \nonumber
\end{align}
The proposed approach is summarized in Algorithm~\ref{alg:algorithm}. We note that our approach can be incorporated in a plug-and-play manner into any neural network-based approach to augment the original loss and to ensure learning diverse features. We also note that although in this paper, we focus only on applying diversity regularizer to a single layer, i.e., the feature layer, our proposed diversity loss, as in  \cite{cogswell2015reducing}, can be applied to multiple layers within the model. 

\begin{algorithm}[tb]
\caption{One epoch of training with WLD-Reg}
\label{alg:algorithm}
\textbf{Model}: Given a neural network $f(\cdot)$ with a feature representation $\phi(\cdot)$, i.e., last intermediate layer. \\
\textbf{Input}: Training Data: $\{\vx_i,y_i\}_{i=1}^{N}$\\
\textbf{Parameters}: $\lambda_1$ and $\lambda_2$ in  \eqref{eq:lossextra} \\
\begin{algorithmic}[1] 
\FOR{ every mini-batch: $\{\vx_i,y_i\}_{i=1}^{m} \in \{\vx_i,y_i\}_{i=1}^{N}  $ }
\STATE Forward pass the inputs $\{\vx_i\}_{i=1}^{m}$ into the model  to obtain the outputs $\{f(\vx_i)\}_{i=1}^{m}$ and the feature representations  $\{\Phi(\vx_i)\}_{i=1}^{m}$ 
\STATE Compute the standard loss $\hat{L}(f) $ (\eqref{eq:loss}).
\STATE Compute the extra loss $\hat{L}_{WLD-Reg}$  (\eqref{eq:lossextra}).
\STATE Compute the total loss $\hat{L}_{aug}(f)$ (\eqref{eq:augloss})
 \STATE Compute the gradient of the total loss and
use it to update the weights of $f$.
\ENDFOR
\STATE \textbf{return} Return $f$.
\end{algorithmic}
\end{algorithm}

Our newly proposed loss function defined in \eqref{eq:augloss} has two terms. The first term is the classic loss function. It computes the loss with respect to the ground-truth. In the back-propagation, this feedback is back-propagated from the last layer to the first layer of the network. Thus, it can be considered as a between-layer feedback, whereas the second term is computed within a layer. From \eqref{eq:augloss}, we can see that our proposed approach can be interpreted as a regularization scheme. However, regularization in deep learning is usually applied directly on the parameters, i.e., weights \cite{goodfellow2016deep,kukavcka2017regularization}, while in our approach a data-dependent additional term is defined over the output maps of the layers.  For a feature layer with $C$ units and a batch size of $m$, the additional computational cost is O($C^2(m+1)$) for Direct variant and O($C^3 + C^2m)$) for both Det and Logdet variants.

\section{Related work}

\textbf{Diversity promoting strategies} have been widely used in ensemble learning \cite{li2012diversity,yu2011diversity}, sampling \cite{biyik2019batch,derezinski2019exact,gartrell2019learning},  energy-based models \cite{laakom2021feature,zhao2016energy}, ranking \cite{gan2020enhancing,yang2019balanced}, pruning by reducing redundancy \cite{he2019filter,kondo2014dynamic,lee2020filter,singh2020leveraging}, and semi-supervised learning \cite{zbontar2021barlow}.  In the deep learning context, various approaches have used diversity as a direct regularizer on top of the weight parameters.  Here, we present a brief overview of these regularizers.  Based on the way diversity is defined, we can group these approaches into two categories. 
The first group considers the regularizers that are based on the pairwise dissimilarity of the components, i.e., the overall set of weights is diverse if every pair of weights is dissimilar. Given the weight vectors $\{\vw_m \}_{m=1}^M$, \cite{yu2011diversity} defines the regularizer as  $\sum_{mn}(1 - \theta_{mn})$, where $\theta_{mn}$ represents the cosine similarity between $\vw_m$ and $\vw_n$. In \cite{bao2013incoherent}, an incoherence score defined as $-\log\left(\frac{1}{M(M-1)} \sum_{mn} \beta |\theta_{mn}|^{\frac{1}{\beta}}\right) $, where $\beta$ is a positive hyperparameter, is proposed. In \cite{xie2015diversifying,xie2016diversity}, $ \text{mean}(\theta_{mn}) - \text{var}(\theta_{mn})$ is used  to regularize Boltzmann machines. The authors theoretically analyzed its effect on the generalization error bounds in \cite{xie2015generalization} and extend it to  kernel space in \cite{xie2017diverse}. The second group of regularizers  considers a more global view of diversity. For example, in \cite{malkin2008ratio,malkin2009multi,xie2017uncorrelation},  a weight regularization based on the determinant of the weights' covariance is proposed based on determinantal point process \cite{kulesza2012determinantal,kwok2012priors}.
 
Unlike the aforementioned methods which promote diversity on the weight level and similar to our method, \cite{cogswell2015reducing,laakom2022reducing} proposed to enforce dissimilarity on the feature map outputs, i.e., on the activations. To this end, they proposed an additional loss based on the pairwise covariance of the activation outputs. Their additional loss, $L_{Decov}$, is defined as the squared sum of the non-diagonal elements of the global covariance matrix $\mC$ of the activations:
\begin{equation} 
L_{Decov} = \frac{1}{2}(||\mC||_F^2 - || \text{diag}(\mC)||_2^2  ),
\end{equation}
where $||.||_F$ is the Frobenius norm. Their approach, Decov, yielded superior empirical performance. However, correlation is highly sensitive to noise \cite{kim2015instability}, as opposite  to the RBF-based distance used in our approach \cite{savas2019impact,haykin2010neural}. Moreover, the $Decov$ approach only captures the pairwise diversity between the components, whereas we propose variants of our approach which consider a global view of diversity. Moreover, based on the cross-covariance, their approach i-s not scale-invariant. In fact, it can be trivially minimized by making all activations in the latent representation  small, which has no effect on the generalization since the model can rescale tiny activations to normal values simply by learning large weights on the next layer.

\section{Experimental results}
\subsection{CIFAR10 \& CIFAR100}
We start by evaluating our proposed diversity approach on two image datasets: CIFAR10 and CIFAR100 \cite{krizhevsky2009learning}.   They contain 60,000 (50,000 train/10,000 test) $32\times32$ images grouped into 10 and 100 distinct categories, respectively. We split the original training set (50,000) into two sets: we use the first 40,000 images as the main training set and the last 10,000 as a validation set for hyperparameters optimization. We use our approach on two state-of-the-art CNNs:
\begin{itemize}
    \item \textbf{ResNext-29-08-16}: we consider the standard ResNext Model \cite{xie2017aggregated} with a 29-layer architecture, a cardinality of 8, and a width of 16. 
    \item \textbf{ResNet50}: we consider the standard ResNet model \cite{he2016deep} with 50 layers.
\end{itemize}    

\begin{table*}[h] 
  \caption{Classification errors of the different approaches on CIFAR10 and CIFAR100 with three different models. Results are averaged over three random seeds.}
  \centering
  \label{cifar}
  \begin{tabular}{lcccc}
   \toprule
          &  \multicolumn{2}{c}{ResNext-29-08-16}   &  \multicolumn{2}{c}{ResNet50}  \\
     \cmidrule(r){2-3}     \cmidrule(r){4-5}    
method  & CIFAR10  &  CIFAR100 &  CIFAR10  &  CIFAR100   \\  
    \midrule
Standard &  6.93 $\pm$  0.10 &   26.73 $\pm$ 0.10  & 8.28 $\pm$  0.41   &  33.39  $\pm$ 0.42 \\  
DeCov &   6.82 $\pm$ 0.15  &  26.70 $\pm$  0.10 &  8.03 $\pm$ 0.11 &    32.26 $\pm$ 0.22 \\  
Ours(Direct) &  \textbf{6.28 $\pm$ 0.11}   &  26.20 $\pm$  0.18   &  7.77 $\pm$ 0.09 &  32.09 $\pm$ 0.11 \\  
Ours(Det)&    6.51 $\pm$ 0.16   &  26.35  $\pm$ 0.23   &   7.75 $\pm$ 0.12  &   32.14 $\pm$ 0.28 \\  
Ours(Logdet) & 6.38 $\pm$ 0.08  & \textbf{25.88 $\pm$ 0.21} &    \textbf{ 7.65  $\pm$    0.10}       &  \textbf{31.99 $\pm$ 0.05} \\  
    \toprule
  \end{tabular}
\end{table*}

We compare against the standard networks\footnote{For the standard approach, the only difference is not using an additional diversity loss. The remaining regularizers, data augmentation, weight decay etc., are all applied as specified per-experiment.} as well as networks trained with the DeCov diversity strategy \cite{cogswell2015reducing}. All the models are trained using stochastic gradient descent (SGD) with  a momentum of 0.9, weight decay of $0.0001$, and a batch size of 128 for 200 epochs. The initial learning rate is set to 0.1 and is then decreased by a factor of 5 after 60, 120, and 160 epochs, respectively. We also adopt a standard data augmentation scheme that is widely used for these two datasets \cite{he2016deep,huang2017densely}. For all models, the additional diversity term is applied on top the last intermediate layer. The penalty coefficients $\lambda_1$ and $\lambda_2$, in \eqref{eq:lossextra}, for our approach and the penalty coefficient of Decov are chosen from $\{0.0001,0.001,0.01,0.1\} $, and $\gamma$ in the radial basis function is chosen from $\{1,10\}$. For each approach, the model with the best validation performance is used in the test phase. We report the average performance over three random seeds. 

Table~\ref{cifar} reports the average top-1 errors of the different approaches with the two basis networks. We note that, compared to the standard approach, employing a diversity strategy consistently boosts the results for all the two models and that our approach consistency outperforms both competing methods (standard and DeCov) in all the experiments. With ResNet50, the three variants of our proposed approach significantly reduce the test errors compared to standard approach over both datasets: $0.51\%-0.63\%$ improvement on CIFAR10 and $1.25\%-1.44\%$ on CIFAR100. 

For CIFAR10, the best performance is achieved by the direct variant and the Logdet variant for ResNext and ResNet models, respectively. For example, with ResNext, our direct variant yields 0.65 boost compared to the standard approach and 0.54 boost compared to DeCov.  For CIFAR100, the best performance is acheived by our Logdet variant for both models. This variant leads to $1.4\%$ and $0.85\%$ boost for ResNet and ResNext, respectively. Overall, our three variants consistently outperform DeCov and standard approach in all testing configurations. 

\subsection{ImageNet}
To further demonstrate the effectiveness of our approach and its ability to boost the performance of state-of-the-art neural networks, we conduct additional image classification experiments on the ImageNet-2012 classification dataset \cite{russakovsky2015imagenet} using four different models: ResNet50 \cite{he2016deep}, Wide-ResNet50 \cite{BMVC2016_87}, ResNeXt50 \cite{xie2017aggregated},  and ResNet101 \cite{he2016deep}. The diversity term is applied on the last intermediate layer, i.e., the global average pooling layer for both DeCov and our method.

For the hyperparameters, we fix $\lambda_1=\lambda_2=0.001$ and $\gamma=10$ for all the different approaches.  The Scope of this paper is feature diversity. However, in this experiment, we also report results with weight diversity approaches. In particular, we compare with the methods in  \cite{yu2011diversity}, \cite{xie2015generalization}, \cite{rodriguez2016regularizing}, and \cite{ayinde2019regularizing}. 

We use the  standard augmentation practice for this dataset as in \cite{zhang2018mixup,huang2017densely,cogswell2015reducing}. All the models are trained with a batch size of 256 for 100 epoch using SGD with Nesterov Momentum of 0.9.  The learning rate is initially set to 0.1 and decreases at epochs 30, 60, 90 by a factor of 10. 
\begin{table*}[h] 
  \caption{Performance of different models with different diversity strategies on ImageNet dataset}
  \label{imagenet_table}
  \centering
  \begin{tabular}{lcccc}
    \toprule
   & ResNet50   &  Wide-ResNet50    &   ResNeXt50     &  ResNet101   \\
\midrule
Standard     &  23.84  &22.42  & 22.70 & 22.33\\
\midrule  
\cite{yu2011diversity}    & 23.87          & 22.48   & 22.57  & 22.23  \\
\cite{ayinde2019regularizing}    &  23.95    & 22.41 & 22.67    & 22.36   \\
\cite{rodriguez2016regularizing}    &  24.23  & 22.70  & 22.80   & 23.10 \\
\cite{xie2015generalization}          & 23.79  & 22.66   & 22.64  & 22.71  \\
\midrule
DeCov   &   23.62        & 22.68  & 22.57  & 22.31 \\
Ours(Direct)  & \textbf{23.24}  & 21.95  &  \textbf{22.25}  & 22.14 \\  
Ours(Det)   &  23.34  & \textbf{21.75}   & 22.44  & \textbf{21.87} \\
Ours(Logdet) &  23.32 & 21.96 & 22.40  & 22.04  \\

\toprule
  \end{tabular}
\end{table*}

Table~\ref{imagenet_table} reports the test errors of the different approaches on ImageNet dataset. As it can be seen, feature diversity (our approach and DeCov) reduces the test error of the model and yields a better performance compared to the standard approach. We note that, as opposite to feature diversity, weight diversity does not always yield performance improvement and it can sometimes hurt generalization. Compared to decov, our three variants consistently reach better performance. 

For ResNet50 and ResNeXt50, the best performance is achieved by our direct variant, yielding more than $0.5\%$ improvement compared to standard approach for both models. For Wide-ResNet50 and ResNet101, our Det variant yields the top performance with over $0.6\%$ boost for Wide-ResNet50. We note that our approach has a small additional time cost. For example for ResNet50, our direct, Det and Logdet variants take only $0.29 \%$,  $0.39\%$, and $0.49\%$ extra training time, respectively.






\subsection{Sensitivity analysis}
\begin{figure*}[h]
\centering
\includegraphics[width=0.32\linewidth]{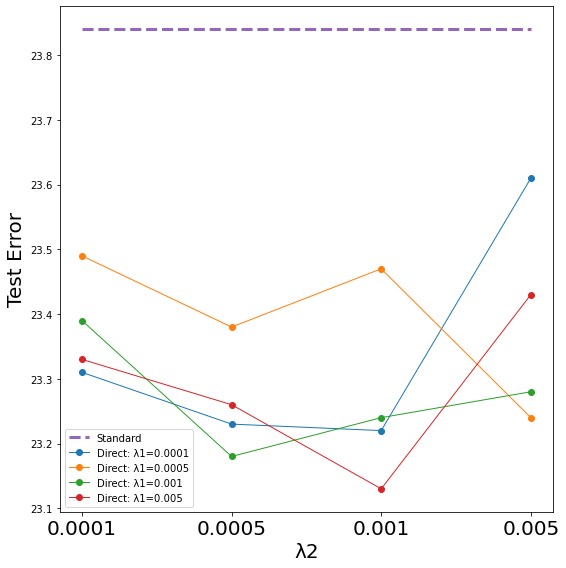}
\includegraphics[width=0.32\linewidth]{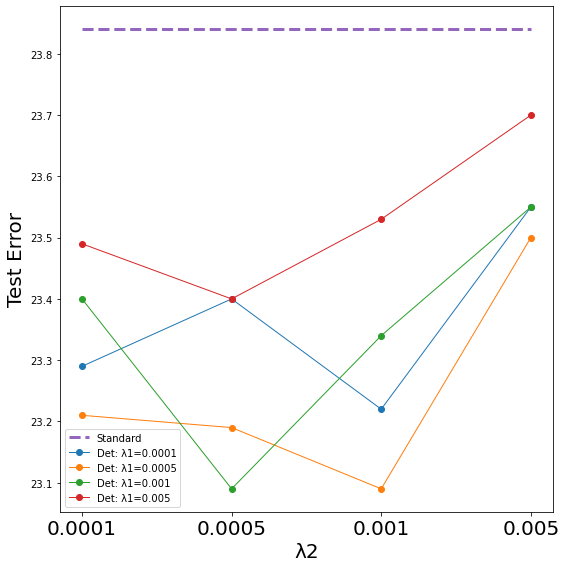}
\includegraphics[width=0.32\linewidth]{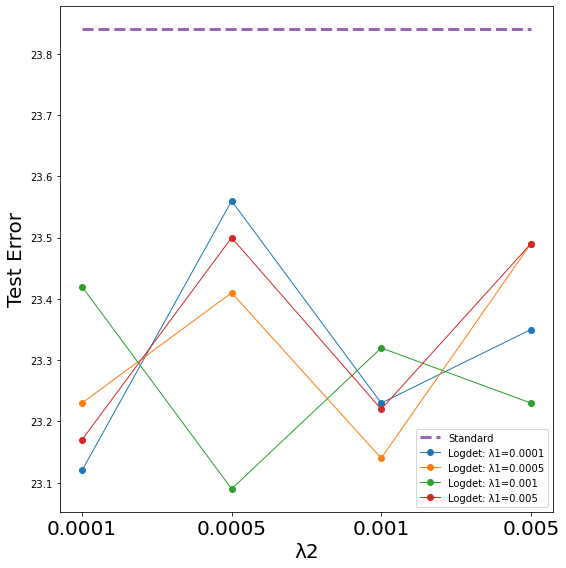}
\includegraphics[width=0.32\linewidth]{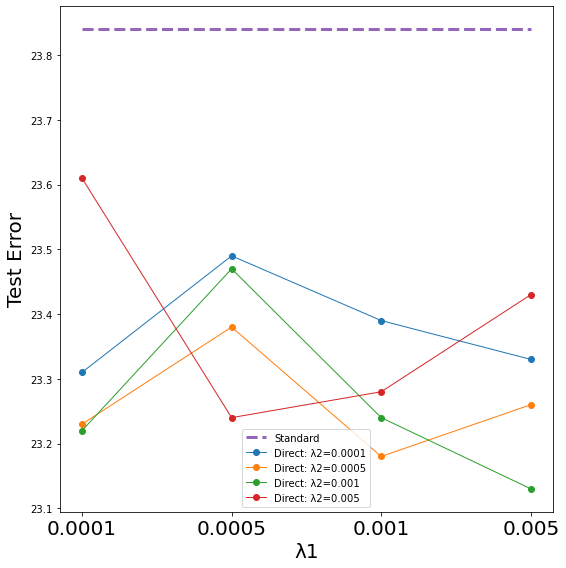}
\includegraphics[width=0.32\linewidth]{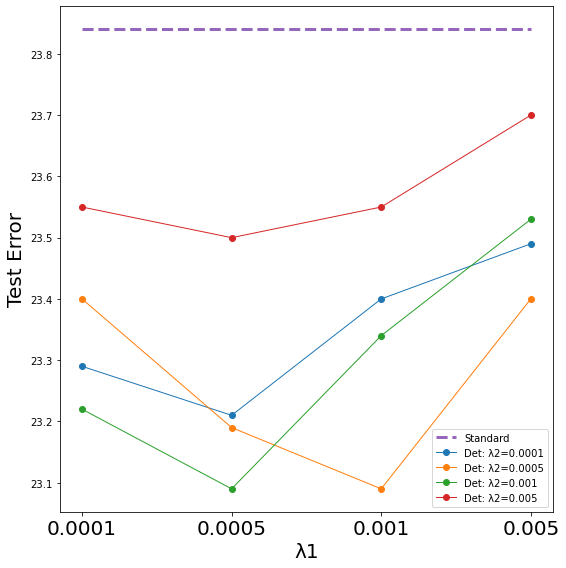}
\includegraphics[width=0.32\linewidth]{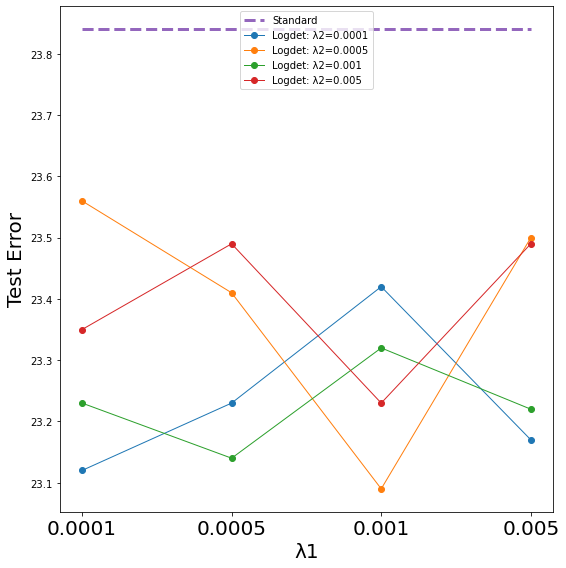}
\caption{Sensitivity analysis of  $\lambda_1$ and $\lambda_2$ on the test error using ResNet50 trained on ImageNet. First row contains experiments with fixed $\lambda_1$ and second row contains experiments with fixed $\lambda_2$. From left to right: our Direct variant, our Det variant and our Logdet variant. $\gamma$ is fixed to 10 in all experiments. }
\label{sensitivity_analysis}
\end{figure*}

To further investigate the effect of the proposed diversity strategy, we conduct a sensitivity analysis using ImageNet on the hyperparameters of our methods:  $\lambda_1$ and $\lambda_2$  which controls the contribution of the global diversity term to the global loss.  We analyse the effect of the two parameters on the final performance of ResNet50 on ImageNet dataset. The analysis is presented in Figure~\ref{sensitivity_analysis}.


As shown in Figure~\ref{sensitivity_analysis},  using a diversity strategy, i.e., three variants of our method, consistently outperform the standard approach and are robust to the hyperparameters. For the Direct variant, the best performance is reached with $\lambda_1=0.005$ and $\lambda_2=0.001$. With this configuration, the model achieve $0.71\%$ improvement compared to the standard approach. For the Det and the Logdet variants, using $\lambda_1=0.001$ and $\lambda_2=0.0005$, the model reaches the lowest error rate ($23.09\%$) corresponding to $0.75\%$ accuracy boost.
Emphasizing diversity and using a high weights ($\lambda_1$ and $\lambda_2$) still lead to better results compared to standard approach but can make the total loss dominated by the diversity term. In general, we recommend using $\lambda_1=\lambda_2=0.001$. However, this depends on the problem at hand. 



\subsection{Feature diversity reduces overfitting}

In \cite{div_theory,cogswell2015reducing}, it has been observed that feature diversity can reduce overfitting.  To study the effect of feature diversity on the generalization gap, in Table \ref{imagenet_table_gap}, we report the final training errors and the generalization gap, i.e., training accuracy - test accuracy for the different feature diversity approaches on ImageNet dataset. 

\begin{table}[h] 
  \caption{Generalization Gap, i.e., training error - test error, of different models with different diversity strategies on ImageNet dataset. * denotes our approach.}
  \label{imagenet_table_gap}
  \centering
  \begin{tabular}{lccccc}
    \toprule
              & ERM   & DeCov    & direct*   & det* & logdet* \\   
\midrule
ResNet50           &  2.87 & 2.70  & \textbf{1.15}  & 1.23 & 1.21 \\
 Wide-ResNet50     & 6.33  & 6.34 &  4.44  & \textbf{4.34} & 4.58 \\
ResNeXt50          & 5.99  & 5.85 &  \textbf{4.41} & 4.59  &  4.48 \\  
ResNet101          & 4.64  & 4.61 &  3.68 &  \textbf{3.38} & 3.71  \\

\toprule
  \end{tabular}
\end{table}

As shown in Table \ref{imagenet_table_gap}, we note that using diversity indeed can reduces overfitting and decreases the empirical generalisation gap of neural networks. The three variants of our approach significantly reduces overfitting for all the four models by more than $1\%$ compared standard and DeCov for all the models.
For example, our Det variant reduces the empirical generalization gap, compared to the standard approach and DeCov, by $~2\%$ for Wide-ResNet model and over $1.2\%$ for the ResNet101 model.

\subsection{MLP-based models}
Beyond CNN models, we also evaluate the performance of our diversity strategy on modern attention-free, multi-layer perceptron (MLP) based models for image classification \cite{tolstikhin2021mlp,liu2021pay,lee2021fnet}.
Such models are known to exhibit high overfitting and require regularization. We evaluate how diversity affects the accuracy of such models on CIFAR10. In particular, we conduct a simple experiment using two models: MLP-Mixer \cite{tolstikhin2021mlp},  gMLP \cite{liu2021pay} with four blocks each.

For the diversity strategies, i.e., ours and Decov, similar to our other experiments, the additional loss has been added on top of the last intermediate layer. The input images are resized to $72\times72$. We use a patch size of $8\times8$ and an embedding dimension of 256. All models are trained for 100 epochs using Adam with learning rate of $0.002$, weight decay with rate $0.0001$, batch size $256$. Standard data augmentation, i.e., random horizontal flip and random zoom with a factor of $20\%$, is used. We use $10\%$ of the training data for validation. We also reduce the learning rate by a factor of 2 if the validation loss does not improve for 5 epochs and use early stopping when the validation loss does not improve for 10 epochs. All experiments are repeated over 10 random seeds and the average results are reported.  

\begin{table}[h] 
        \caption{Classification errors of modern
        MLP-based approaches on CIFAR10. Results are averaged over ten random seeds.}
      \label{all_MLP}
  \centering
  \begin{tabular}{lcc}
    \toprule
  & \multicolumn{1}{c}{MLP-Mixer}  & \multicolumn{1}{c}{ gMLP}  \\
    \midrule
Standard  & 23.93  & 22.26 \\ 
DeCov  & 24.10 &  22.00  \\ 
Ours(Direct) & 22.78 & 21.95  \\ 
Ours(Det)   & \textbf{22.66}  &  21.62  \\ 
 Ours(Logdet)   &  22.84 & \textbf{21.56}  \\ 
    \toprule
  \end{tabular}
\end{table}

The results in Table \ref{all_MLP} show that employing a diversity strategy can indeed improve the performance of these models, thanks to its ability to help learn rich and robust representation of the input. Our proposed approach consistently outperforms the competing methods for both the MLP-Mixer and gMLP. For example, our direct variant leads to $1.15\%$ and $0.3\%$ boost for MLP-Mixer and gMLP, respectively.

For the MLP-mixer, the top performance is achieved by the Det variant of our approach reducing the error rates by $1.27\%$ and $1.44\%$ compared to the standard approach and DeCov, respectively. For the gMLP model, the top performance is achieved by the Logdet variant of our approach boosting the results by $0.7\%$ and $0.44\%$ compared to the standard approach and DeCov, respectively.



\subsection{Learning in the presence of label noise}

\begin{table*}[h] 
  \caption{Classification errors of ResNet50 using different diversity strategies on CIFAR10 and CIFAR100 datasets with different label noise ratios.  Results are averaged over three random seeds.}
  \label{noise_label}
  \centering
  \begin{tabular}{lllll}
    \toprule
  & \multicolumn{2}{c}{20\% label noise}   &   \multicolumn{2}{c}{40\% label noise}      \\
     \cmidrule(r){2-3} \cmidrule(r){4-5}
Method     &  CIFAR10 &  CIFAR100 &  CIFAR10   &  CIFAR100 \\
    \midrule
Standard     & 14.38 $\pm$ 0.29  & 45.11  $\pm$  0.52 & 19.40 $\pm$ 0.80    & 48.81 $\pm$ 0.57   \\ 
DeCov        &  13.75 $\pm$  0.19  &  41.93 $\pm$ 0.40 & 17.60  $\pm$ 0.66 &  48.23  $\pm$ 0.48 \\
Ours(Direct)  & 13.31 $\pm$ 0.40   & 40.10 $\pm$ 0.31  &  \textbf{16.96 $\pm$ 0.32}   &  46.73 $\pm$  0.23 \\ 
Ours(Det)     & 13.21 $\pm$ 0.21   &  40.35 $\pm$ 0.31 & 17.49 $\pm$   0.04  & 46.93 $\pm$ 0.62 \\ 
Ours(Logdet)  & \textbf{13.01 $\pm$ 0.40}  & \textbf{39.97 $\pm$ 0.19}  &  17.24 $\pm$ 0.31   & \textbf{46.52 $\pm$ 0.22} \\ 
    \toprule
  \end{tabular}
\end{table*}

To further demonstrate the usefulness of promoting diversity, we test the robustness of our approach in the  presence of label noise. In such situations, standard neural network tend to overfit to the noisy samples and not generalize well to the test set. Enforcing diversity can lead to better and richer representations attenuating the effect of noise. To show this, we performed additional experiments with label noise (20\% and 40\%) on CIFAR10 and CIFAR100 using ResNet50. We use the same training protocol used for the original CIFAR10 and CIFAR100: all models are trained using SGD with a momentum of 0.9, weight decay of $0.0001$, and a batch size of 128 for 200 epochs. The initial learning rate is set to 0.1 and is then decreased by a factor of 5 after 60, 120, and 160 epochs, respectively. We also adopt a standard data augmentation scheme that is widely used for these two datasets \cite{he2016deep,huang2017densely}. For all models, the additional diversity term is applied on top the last intermediate layer.  For the hyperparameters: The loss weights is chosen from $\{0.0001,0.001,0.01,0.1\} $ for both our approach ($\lambda_1$ and $\lambda_2$) and Decov and $\gamma$ in the radial basis function is chosen from $\{1,10\}$. For each approach, the model with the best validation performance is used in the test phase. The average errors over three random seed are reported.

The results are reported in Table \ref{noise_label}. As it can be seen, in the presence of noise, the gap between the standard approach and diversity (Decov and ours) increases. For example, our Logdet variant boosts the results by 1.91\% and 2.29\% on CIFAR10 and  CIFAR100 with 40\% noise, respectively.

\section{Conclusions }
In this paper, we proposed a new approach to encourage  `diversification' of the layer-wise feature map outputs in neural networks. The main motivation is that by promoting within-layer activation diversity, units within the same layer learn to capture mutually distinct patterns. We proposed an additional loss term that can be added on top of any fully-connected layer. This term complements the traditional `between-layer' feedback with an additional `within-layer' feedback encouraging  diversity of the activations. Extensive experimental results showing that such a strategy can indeed improve the performance of different state-of-the-art networks across different datasets and different tasks, i.e., image classification, and label noise. We are confident that these results will spark further research in diversity-based approaches to improve the performance of neural networks.

\bibliography{aaai22} 

\end{document}